\documentclass[letterpaper]{article} 
\usepackage{aaai25}  
\usepackage{times}  
\usepackage{helvet}  
\usepackage{courier}  
\usepackage[hyphens]{url}  
\usepackage{graphicx} 
\usepackage{natbib}  
\usepackage{caption} 
\usepackage{tcolorbox}
\tcbuselibrary{listingsutf8}
\usepackage{listings}
\usepackage{tcolorbox}
\tcbuselibrary{listingsutf8}
\usepackage{tcolorbox}
\usepackage{amsfonts}
\usepackage{listings}
\usepackage{xcolor}
\tcbuselibrary{listings,skins}
\usepackage{amsmath}
\usepackage{booktabs}   
\usepackage{multirow} 

\usepackage{algorithm}
\usepackage{algorithmic}

\usepackage{newfloat}
\usepackage{listings}
\DeclareCaptionStyle{ruled}{labelfont=normalfont,labelsep=colon,strut=off} 
\floatstyle{ruled}
\newfloat{listing}{tb}{lst}{}
\floatname{listing}{Listing}
\title{Assessing the Geolocation Capabilities, Limitations and Societal Risks of Generative Vision-Language Models}
\author{
    O. Grainge\textsuperscript{\rm 1}\equalcontrib, S. Waheed\textsuperscript{\rm 1}\equalcontrib, J. Stilgoe\textsuperscript{\rm 2}, M. Milford\textsuperscript{\rm 3}, S. Ehsan\textsuperscript{\rm 1, 4} \\
}
\affiliations{
    \textsuperscript{\rm 1}University of Southampton, 
    \textsuperscript{\rm 2}University College London, 
    \textsuperscript{\rm 3}Queensland University of Technology, 
    \textsuperscript{\rm 4}University of Essex\\

    oeg1n18@soton.ac.uk
    s.waheed@soton.ac.uk 
    j.stilgoe@ucl.ac.uk  michael.milford@qut.edu.au
    s.ehsan@soton.ac.uk

}

\begin{document}

\maketitle

\begin{abstract}
Geo-localization is the task of identifying the location of an image using visual cues alone. It has beneficial applications, such as improving disaster response, enhancing navigation, and geography education. Recently, Vision-Language Models (VLMs) are increasingly demonstrating capabilities as accurate image geo-locators. This brings significant privacy risks, including those related to stalking and surveillance, considering the widespread uses of AI models and sharing of photos on social media. The precision of these models is likely to improve in the future. Despite these risks, there is little work on systematically evaluating the geolocation precision of Generative VLMs, their limits and potential for unintended inferences. To bridge this gap, we conduct a comprehensive assessment of the geolocation capabilities of 25 state-of-the-art VLMs on four benchmark image datasets captured in diverse environments. Our results offer insight into the internal reasoning of VLMs and highlight their strengths, limitations, and potential societal risks. Our findings indicate that current VLMs perform poorly on generic street-level images yet achieve notably high accuracy (61\%) on images resembling social media content, raising significant and urgent privacy concerns.


\end{abstract}

\begin{figure}[t!]
\centering
\includegraphics[width=1.0\linewidth]{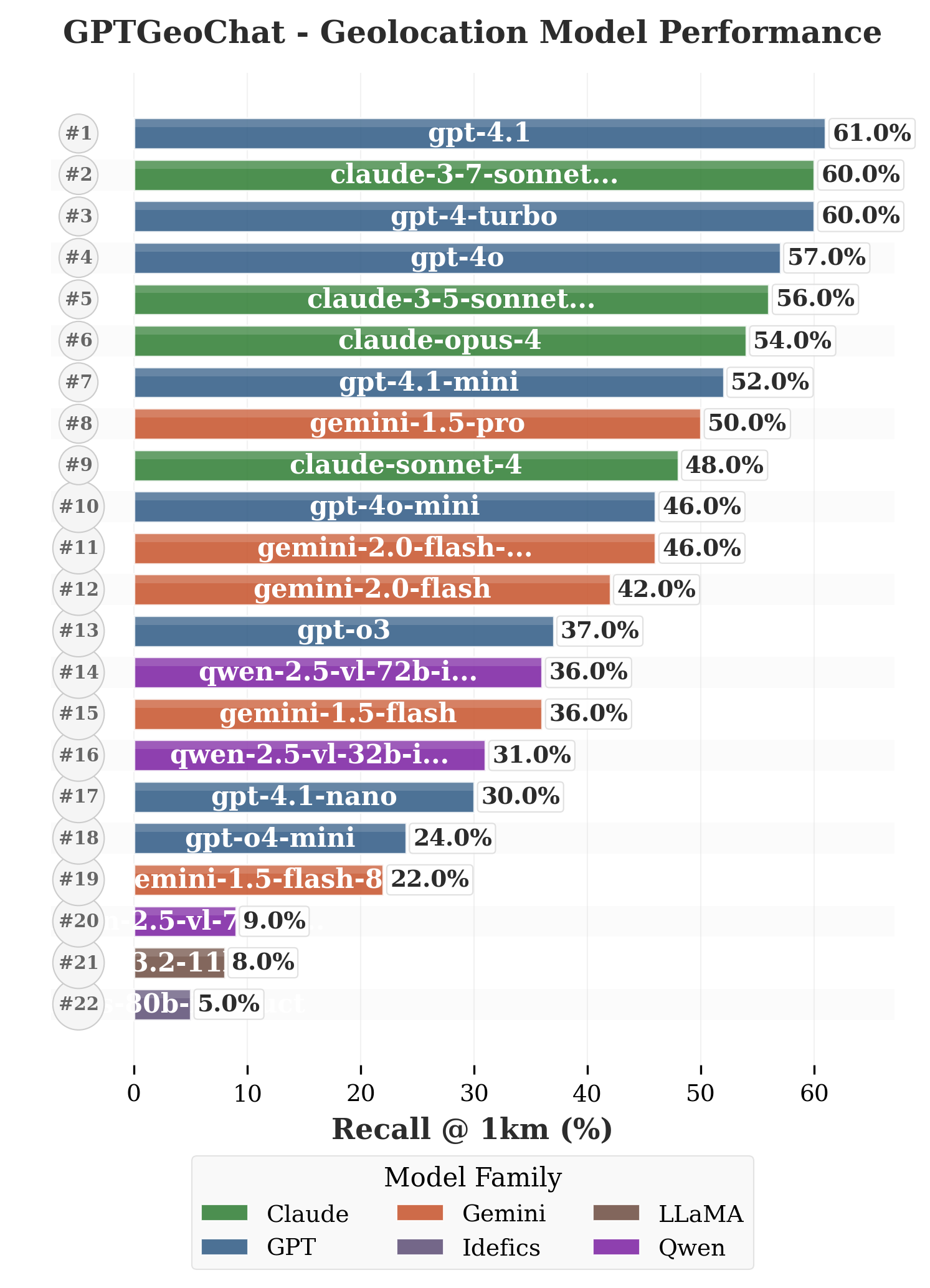}
\caption{Leaderboard of 22 open-source, open-weight and closed-source vision–language models (VLMs) on the GPTGeoChat dataset \cite{mendes2024granular}. Models are ranked by Recall@1km, which is the percentage of predictions within 1 km of the ground truth. Models with 0\% recall or a failure rate above 90\% are excluded.}
\label{fig:leaderboard}
\end{figure}

\section{Introduction} \label{sec:introduction}
Geo-localization, the task of identifying the geographic location of an image using only its visual content, is an important capability for applications such as disaster response, navigation, and geographic education. Traditional approaches have primarily relied on supervised learning with large geotagged datasets or on large-scale image retrieval methods. Recent advances in Vision-Language Models (VLMs) have revealed that these models possess emergent geo-localization abilities, even without task-specific training. Early work in this area focused on models such as CLIP \cite{geoclip, haas2023learninggeneralizedzeroshotlearners, 10.1145/3557918.3565868}, but more recent efforts have applied generative VLMs to this task \cite{Zhou_2024, mendes2024granular, waheed2025imagebasedgeolocalizationroboticsblackbox}. These models have shown surprisingly strong performance on geo-localization benchmarks, often matching or exceeding the performance of specialized geo-localization models \cite{mendes2024granular}.

The rapid improvement of VLM-based geo-localization offers significant opportunities but also introduces serious risks. The widespread availability and multipurpose nature of VLMs make their use difficult to regulate. Precise geo-localization from publicly shared images can be exploited for stalking, surveillance or discrimination. As the performance and accessibility of VLMs continue to improve, developers and users of VLMs will look for new opportunities to generate value from geolocation. In the absence of strong regulation, risks will grow. We see risk in sociotechnical terms \cite{doi:10.1126/science.adi8982}. Risk and safety are not just functions of the capabilities of AI models, but of the societal context in which these models sit, and in which they are developed, used and regulated. From this view, privacy is not just about protecting the rights of individuals; it also functions as an important check on the power of those developing and deploying technologies with potential surveillance uses \cite{solove2025privacy}. A rigorous evaluation of the capabilities and limitations of these models is therefore essential for responsible deployment and for informing policymakers about potential misuse and the need for safeguards.

Despite these concerns, there is limited systematic research on the geo-localization performance of generative VLMs. While prior work has examined data leakage and memorization in large language models (LLMs) \cite{Carlini2020ExtractingTD, lm-memorization}, few studies have investigated how generative VLMs perform in geo-localization across diverse, real-world contexts \cite{waheed2025imagebasedgeolocalizationroboticsblackbox}. Moreover, details of important factors such as differences between model families, the effect of model size, and failure cases have largely been overlooked.

To bridge this gap, we present a comprehensive black-box evaluation of 25 state-of-the-art generative VLMs, including open-source, open-weight and closed-source systems. We evaluate these models on four benchmark datasets that collectively capture a broad range of scenarios, including globally distributed images (im2gps \cite{hays2008im2gps} and im2gps3k \cite{vo2017revisitingim2gpsdeeplearning}), urban street-level imagery (OSV5 \cite{osv5m}), and data curated to resemble images commonly posted on social media (GPTGeoChat \cite{mendes2024granular}). In addition to measuring accuracy, we also account for cases where models fail to respond. 

Our results provide the first large-scale comparative study of generative VLMs in geo-localization and highlight both their performance potential and the privacy risks that arise from their growing deployment in real-world applications. Our key contributions are:
\begin{itemize}
    \item We benchmark 25 state-of-the-art generative VLMs on four diverse datasets, offering a detailed comparison of their strengths and limitations.
    \item We analyze the broader societal and ethical risks associated with VLM-based geo-localization, offering insights for future research and policy.
\end{itemize}

The remainder of this paper is organized as follows. The Related Work section summarizes prior research. Experimental Setup details the models, datasets, evaluation metrics, and methodology. Results and Analysis presents our experimental results and a comprehensive performance analysis. Discussion and Implications examines the societal impact and broader implications of our findings. Finally, Conclusion summarizes the study and outlines future research directions.

\section{Related Work}\label{sec:related_works}
This section covers prior research relevant to our work, covering both technical approaches to geo-localization and emerging concerns around privacy and societal implications.
\subsection{Geo-localization and Vision-Language Models}
Traditional geo-localization research \cite{hays2008im2gps} treats localization as a nearest-neighbor problem over a large-scale geotagged image database. This approach works well for landmark recognition \cite{arandjelovic2016netvlad, weyand2020google, kendall2015posenet, wang2020online} but faces scalability challenges and perceptual aliasing at global levels \cite{muller2018geolocation, regmi2019bridging}. To overcome these limitations, classification-based methods \cite{weyand2016planet, muller2018geolocation, kordopatis2021leveraging, izbicki2020exploiting} partition the Earth into discrete geo-cells,  enabling broader global reasoning but introduce trade-offs between spatial resolution and class imbalance \cite{kordopatis2021leveraging, haas2024pigeonpredictingimagegeolocations}.

Recent advances in Vision-Language Models (VLMs) have introduced a new paradigm for geo-localization. Models like CLIP \cite{radford2021learningtransferablevisualmodels} introduce semantic reasoning and contextual understanding to feature representations \cite{10.1145/3557918.3565868, haas2024pigeonpredictingimagegeolocations, luo2022g3geolocationguidebookgrounding, Zhou_2024}. More recently, generative VLMs such as GPT-4v \cite{achiam2023gpt} have demonstrated the ability to infer geographic locations directly from images, sometimes surpassing specialized geo-localization models in accuracy \cite{mendes2024granular, waheed2025imagebasedgeolocalizationroboticsblackbox}. 

Despite these advances, existing evaluations remain limited. Most prior work examines only a handful of models, leaving critical key factors such as model family, scale, training paradigm and failure cases underexplored. These gaps highlight the need for a systematic, large-scale image-based geo-localization benchmarking effort focused on generative VLMs.

\begin{table*}[t]
\centering
\caption{Vision-Language Model Performance Across Geolocation Datasets.} 
\label{tab:geolocation_performance}
\footnotesize
\setlength{\tabcolsep}{3pt}
{\renewcommand{\arraystretch}{1}
\begin{tabular}{@{}lccccccc@{}}
\toprule
\multirow{2}{*}{Model} & \multicolumn{3}{c}{R@1km (\%)} & R@100km (\%) & \multicolumn{3}{c}{Avg. across datasets} \\
 & GPTGeoChat & Im2GPS & Im2GPS3k & OSV5 & City Acc (\%) & Country Acc (\%) & Failure Rate (\%) \\
\midrule
Claude-3.5-S & 56.0 & 17.3 & 12.8 & 3.4 & 31.7 & 61.1 & 0.3 \\
Claude-3.7-S & 60.0 & 19.0 & 12.5 & 4.0 & 32.0 & 62.3 & 0.4 \\
Claude-Opus & 54.0 & 15.2 & 11.3 & 1.0 & 29.1 & 58.5 & 0.3 \\
Claude-S4 & 48.0 & 14.8 & 9.8 & 0.8 & 27.2 & 57.5 & 0.0 \\
\midrule
GPT-4-T & 60.0 & 21.9 & 16.1 & 4.0 & 35.1 & 65.8 & 0.1 \\
GPT-4.1 & 61.0 & 23.2 & 19.1 & 11.6 & 36.6 & 76.5 & 0.0 \\
GPT-4.1-M & 52.0 & 16.5 & 12.1 & 4.6 & 30.3 & 67.5 & 0.0 \\
GPT-4.1-N & 30.0 & 5.9 & 4.7 & 0.2 & 18.0 & 47.3 & 0.7 \\
GPT-4o & 57.0 & 20.7 & 16.3 & 6.2 & 38.0 & 75.8 & 11.0 \\
GPT-4o-M & 46.0 & 14.8 & 12.3 & 1.2 & 32.9 & 70.8 & 21.7 \\
GPT-o3 & 37.0 & 16.9 & 11.8 & 6.0 & 41.3 & 77.5 & 33.1 \\
GPT-o4-M & 24.0 & 10.5 & 7.9 & 1.2 & 37.8 & 72.0 & 61.4 \\
\midrule
Gemini-1.5-F & 36.0 & 12.7 & 10.9 & 9.4 & 31.0 & 72.4 & 0.0 \\
Gemini-1.5-8B & 22.0 & 6.8 & 6.6 & 5.0 & 26.3 & 70.9 & 1.4 \\
Gemini-1.5-P & 50.0 & 18.6 & 14.8 & 9.6 & 33.0 & 74.1 & 0.0 \\
Gemini-2.0-F & 42.0 & 16.5 & 10.0 & 4.4 & 32.4 & 70.3 & 7.5 \\
Gemini-2.0-FL & 46.0 & 16.0 & 12.5 & 7.8 & 32.8 & 73.9 & 0.2 \\
\midrule
Gemma-3-12B & 0.0 & 0.0 & 0.2 & 0.0 & 2.3 & 29.5 & 0.0 \\
Gemma-3-27B-it & 0.0 & 0.0 & 0.2 & 0.0 & 2.3 & 29.5 & 0.0 \\
\midrule
LLaMA-3.2-11B & 8.0 & 4.2 & 1.8 & 0.0 & 19.1 & 45.3 & 22.6 \\
LLaMA-3.2-90B & 0.0 & 14.8 & 8.7 & 1.6 & 12.8 & 38.1 & 6.8 \\
\midrule
Qwen-2.5-72B & 36.0 & 10.1 & 6.9 & 2.0 & 29.0 & 65.7 & 0.0 \\
Qwen-2.5-32B & 31.0 & 10.5 & 7.1 & 2.0 & 25.0 & 56.6 & 0.0 \\
Qwen-2.5-7B & 9.0 & 4.2 & 4.3 & 0.8 & 27.3 & 59.0 & 4.3 \\
\midrule
Idefics-80B & 5.0 & 2.1 & 1.4 & 0.0 & 20.2 & 70.1 & 75.5 \\
\bottomrule
\end{tabular}
}
\end{table*}

\subsection{Privacy and Societal Risks}
The sensitivity of location data has long been a central concern in privacy research, as even seemingly innocuous information can reveal sensitive personal details. The \textit{privacy paradox} highlights a persistent disconnect between individuals’ stated concerns and their behavior: while people claim to value privacy, they often disclose personal data freely and fail to take protective measures \cite{GERBER2018226, IJoC4655, KOKOLAKIS2017122}. This gap between awareness and action is particularly dangerous for location data, which can reveal intimate aspects of daily life, including routines, habits, and social connections.

Prior studies have shown that even limited location traces can be used to reconstruct or predict a person’s movements with surprising accuracy \cite{9861098}. While earlier research primarily focused on structured data sources, such as GPS logs or check-ins \cite{10.1007/s007790200035, Ashbrook, 8029302}, advances in VLMs have introduced new risks. Visual content shared online, such as social or travel photos, can now be reverse-engineered to reveal precise geographic locations \cite{mendes2024granular}. This capability raises risks not only for individual privacy but also for public safety by enabling stalking, targeted harassment, or even large-scale surveillance.

\section{Experimental Setup}\label{sec:Experimental_Setup}

This section details the experimental framework used to evaluate the geolocation capabilities of generative Vision-Language Models (VLMs). We describe the models under study, the datasets used for evaluation, the prompting and data extraction pipeline, and the metrics applied for quantitative and categorical performance analysis.

\subsection{Vision-Language Models}

We benchmark \textbf{25 generative Vision-Language Models (VLMs)}, spanning seven distinct architecture families and parameter scales ranging from 4 billion to over 90 billion parameters (with some closed-source models having undisclosed parameter counts). To facilitate principled and reproducible analysis, we categorize these models into three transparency-based licensing regimes, reflecting their availability and auditability:

\begin{itemize}
\item \textbf{Closed-source:} Access is restricted to proprietary APIs, with undisclosed model weights, training datasets, and data-filtering heuristics.
\item \textbf{Open-weight:} Model checkpoints and inference code are publicly available, but training data, preprocessing pipelines, and optimization details remain undisclosed.
\item \textbf{Open-source:} Full release of model weights, training code, inference code, and comprehensive documentation of the data processing pipeline, enabling complete reproducibility.
\end{itemize}

\paragraph{Closed-source models.}
We include OpenAI’s \textbf{GPT-o3, GPT-o4-mini, GPT-4-Turbo, GPT-4o, and GPT-4.1}; Anthropic’s \textbf{Claude-3.5, Claude-3.7, Claude-4.0, and Claude-Opus}; and Google’s \textbf{Gemini-1.5 and Gemini-2.0}. While detailed architectures of these systems are proprietary, literature and official documentation indicate they typically employ large-scale transformer architectures, often leveraging mixture-of-experts (MoE) techniques and proprietary visual embedding modules. Prior evaluations \cite{openai2024gpt4technicalreport, hochmair2024correctness} confirm state-of-the-art performance in multimodal reasoning tasks, particularly in vision-language reasoning and spatial inference. The inclusion of these models is motivated both by their empirical performance and their potential implications in realistic high-stakes geospatial misuse scenarios.

\paragraph{Open-weight models.}
We evaluate Meta’s \textbf{LLaMA-3 Vision} models (11B, 90B) \cite{dubey2024llama}, Alibaba’s \textbf{Qwen-2.5-VL} series (7B, 32B, 72B) \cite{Qwen2.5-VL}, and Google’s \textbf{Gemma-3} models (4B, 12B) \cite{gemmateam2025gemma3technicalreport}. These models publicly release checkpoints and inference code but obscure critical pre-training details, including visual dataset composition, data augmentation, filtering heuristics, and optimization procedures. Architecturally, these models commonly use unified transformer backbones with learned multimodal embeddings or fusion modules, integrating vision transformer encoders adapted to visual information before language modeling. These open-weight models represent the frontier of publicly accessible checkpoints with strong cross-domain generalization.

\paragraph{Open-source models.}
We benchmark Hugging Face’s \textbf{IDEFICS} model (80B), which uniquely provides not only checkpoints and inference code but also full training scripts, data preprocessing pipelines, and filtering heuristics under a permissive open-source license \cite{laurenccon2024matters}. These are the only models in our benchmark allowing full replication from raw input through to final training, providing an important transparency-driven baseline for evaluating reproducibility and auditability.

\paragraph{Selection rationale.}
Our model selection intentionally spans transparency regimes, architectural diversity, and parameter scale to systematically examine how each factor influences model performance and behavior in geolocation-related tasks. We explicitly include models known to have specialized multimodal capabilities (e.g., Claude-Opus and GPT-4o), which reportedly leverage advanced spatial attention mechanisms and long-context reasoning. By including multiple models in each transparency category, we aim to disentangle effects stemming from transparency, modality integration strategies, and model capacity, allowing comprehensive evaluation of risks associated with advanced proprietary systems and publicly available alternatives.

All models are evaluated through a unified inference pipeline to ensure comparability and eliminate confounding effects from interface-specific preprocessing or postprocessing strategies.

\subsection{Datasets}

We evaluate on four benchmark datasets, each comprising geo-tagged images with precise latitude and longitude annotations. All explicit metadata is removed to ensure predictions are based solely on visual content.

\begin{itemize}
    \item \textbf{IM2GPS} and \textbf{IM2GPS3k:} Global Flickr-sourced datasets with diverse natural, rural, and urban outdoor imagery. im2gps includes 237 images, while im2gps3k scales this to 3,000.
    
    \item \textbf{GPTGeoChat:} A curated dataset of 1,000 social-media-like images, approximately evenly split between U.S. and international locations. Roughly 85\% of the images include embedded text, reflecting realistic multimodal content.
    
    \item \textbf{OSV5:} A street-view dataset spanning 225 countries and 70,000+ cities. For tractability, we randomly sample 500 images, ensuring broad geographic representation, including non-Western and rural areas.
\end{itemize}

\subsection{Prompting and Output Extraction}

Each image is presented to the VLM with a standardized prompt designed to elicit structured predictions across multiple geographic levels: country, city, neighborhood, and exact location. The prompt requires complete, reasoned responses in a valid JSON format.

\paragraph{Standardized Prompt:}
\textit{Please provide your speculative guess for the location of the image at the country, city, neighborhood, and exact location levels... [prompt omitted for brevity]}.

\begin{tcolorbox}[colback=gray!3!white, colframe=gray!60!black, title=Expected JSON Output Format, fonttitle=\bfseries, coltitle=black, colbacktitle=gray!15!white, boxrule=0.8pt, arc=0pt, left=0mm, right=0mm, top=2mm, bottom=2mm, enhanced]
\begin{lstlisting}[basicstyle=\ttfamily\footnotesize, breaklines=true, columns=flexible, showstringspaces=false, numbers=none, frame=none, backgroundcolor=\color{white}]
{
  "rationale": "Country: I chose United States because...",
  "country": "United States",
  "city": "New York City",
  "neighborhood": "Manhattan",
  "exact_location_name": "Empire State Building",
  "latitude": "40.748817",
  "longitude": "-73.985428"
}
\end{lstlisting}

\end{tcolorbox}

\paragraph{Postprocessing:}
We extract structured predictions using regex-based parsing to accommodate minor generation errors. For latitude and longitude, we match coordinates using: \verb|r"-?\d{1,3}.\d+"|. This ensures robust recovery even when the model deviates slightly from strict JSON formatting.

\subsection{Evaluation Metrics}
Model outputs are evaluated using both continuous geospatial metrics and discrete classification scores.

\paragraph{Haversine Distance.}
The core metric used in our quantitative evaluation is the haversine distance between predicted $(\hat{\phi}, \hat{\lambda})$ and ground-truth $(\phi, \lambda)$ coordinates:
\begin{equation}
d = 2r \arcsin \left( \sqrt{ \sin^2\left( \frac{\Delta \phi}{2} \right) + \cos(\phi)\cos(\hat{\phi})\sin^2\left( \frac{\Delta \lambda}{2} \right) } \right),
\end{equation}
where $r = 6{,}371$ km is Earth's radius. Here, $\phi$ and $\lambda$ denote latitude and longitude, respectively, given in radians. $\Delta\phi = \hat{\phi} - \phi$ and $\Delta\lambda = \hat{\lambda} - \lambda$ represent the differences between the predicted and ground-truth latitude and longitude.

\begin{figure*}[t!]
\centering
\includegraphics[width=\textwidth]{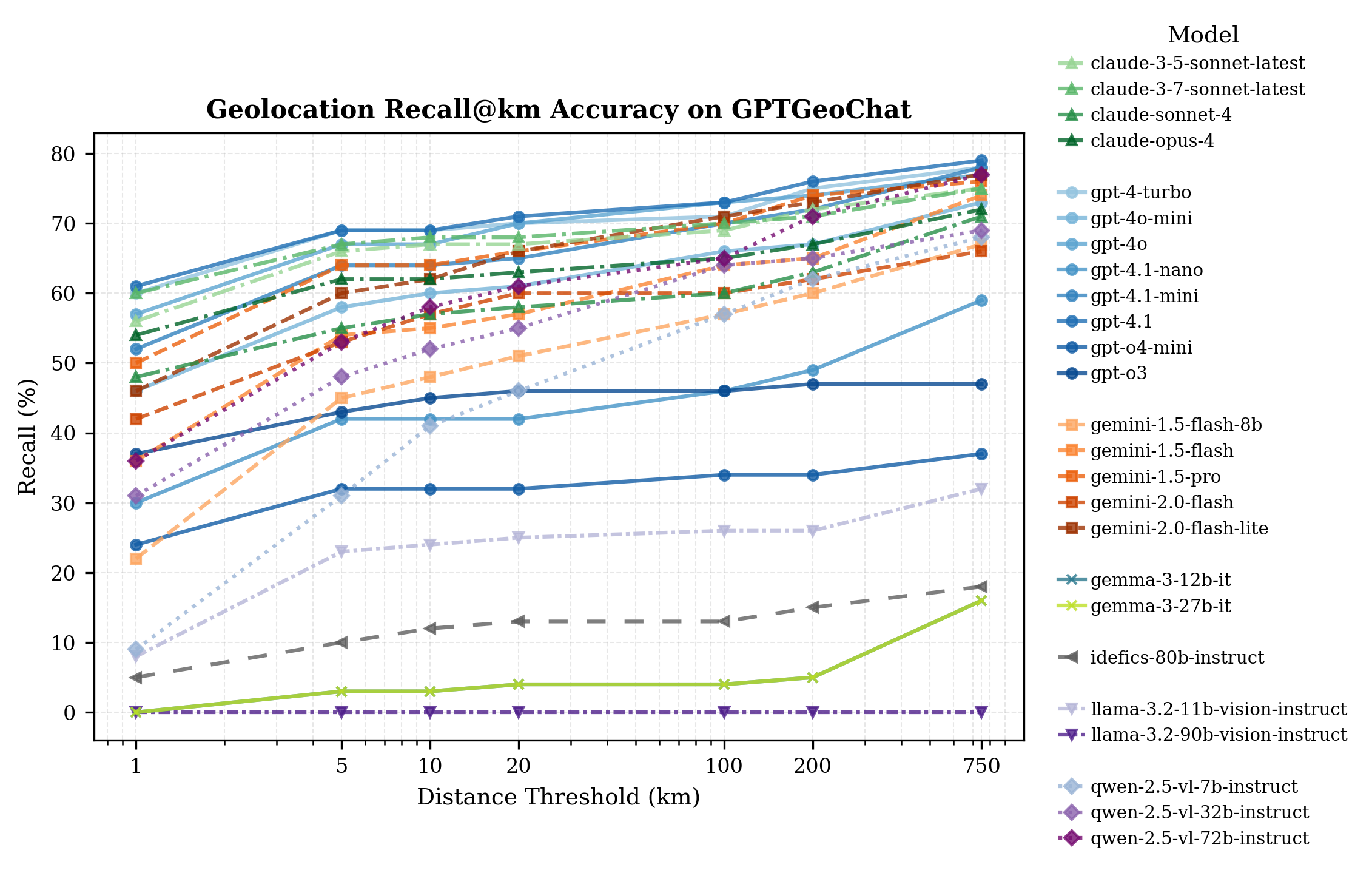}
\caption{Recall@km performance of 25 open-weight, and open- and closed-source generative VLMs evaluated on the GPTGeoChat dataset.}
\label{fig:recall@km}
\end{figure*}

\paragraph{Recall@Nkm.}
We compute Recall@Nkm, the proportion of predictions within $N$ km haversine distance of ground truth:

\begin{equation}
\text{Recall@}N = \frac{1}{M} \sum_{i=1}^{M} \mathbb{I}[d_i \leq N],
\end{equation}

reporting Recall@1, 25, 200, and 750 km to cover street-, city-, regional-, and country-level granularity.

\paragraph{Administrative Accuracy.}
We report classification accuracy at the country and city levels:

\begin{equation}
\text{Accuracy}_{\text{country/city}} = \frac{1}{M} \sum_{i=1}^{M} \mathbb{I}[\hat{c}_i = c_i]
\end{equation}

\begin{figure*}[t!]
\centering
\includegraphics[width=\textwidth]{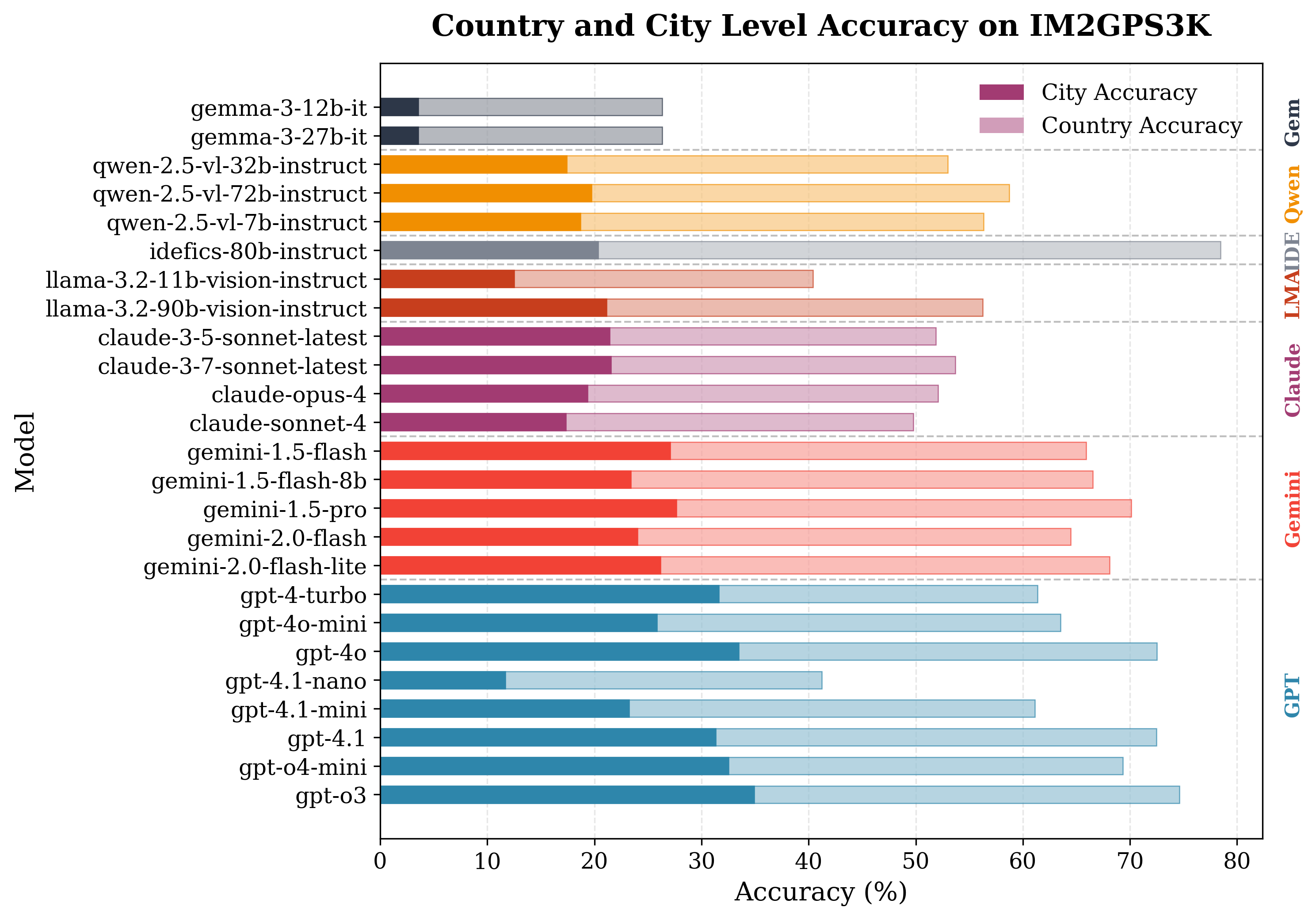}
\caption{Categorical accuracy (city- and country-level) of 25 generative VLMs evaluated on the IM2GPS3k dataset.}
\label{fig:im2gps3k}
\end{figure*}

All labels are normalized using services like GeoNames or OpenStreetMap to resolve spelling variants and naming inconsistencies.

This mixed evaluation approach quantifies both spatial proximity and semantic localization fidelity, offering a comprehensive assessment of model geolocation capabilities.

\section{Results and Analysis}

This section provides a comprehensive analysis of the geolocation capabilities of the evaluated Vision-Language Models (VLMs). Our analysis is structured into three subsections: \textit{Quantitative Localization Accuracy}, which focuses on distance-based Recall@km metrics; \textit{Administrative Region Accuracy}, where we evaluate categorical predictions at city and country levels; and \textit{Failure Analysis}, where we evaluate the ability of the models to provide a geolocation estimation.

\subsection{Quantitative Localization Accuracy}

Table \ref{tab:geolocation_performance} summarizes quantitative results across the four benchmark datasets, with figure \ref{fig:recall@km} showing the Recall@km accuracy across distances on the GPTGeoChat dataset. Overall, GPT-4.1 consistently emerges as the strongest performer, achieving the highest recall across all evaluated distance thresholds (1km, 25km, 200km, and 750km). Specifically, GPT-4.1 attains a remarkable Recall@1km of 61\% on GPTGeoChat, significantly outperforming other closed-source models such as Claude-3.7-Sonnet (60\%) and Gemini-1.5-Pro (50\%).

In comparison, open-weight models exhibit a notable gap in performance. The best-performing open-weight model, Qwen-2.5-72B, achieves a Recall@1km of 36\% on GPTGeoChat, indicating less precise geospatial understanding. Meanwhile, fully open-source models (IDEFICS) show limited geolocation capabilities, with a maximum Recall@1km of only 5\%. This modest but measurable performance indicates that existing data mixtures used in its training, such as Wikipedia, OBELICS, and LAION, provide some foundational support for geolocation tasks albeit with relatively low precision.

Performance varies considerably depending on dataset characteristics. Closed-source models perform exceptionally well on GPTGeoChat, which closely mimics social media imagery containing distinctive visual and textual cues. In contrast, performance drops substantially for datasets such as OSV5, characterized by broader geographic diversity and fewer recognizable visual landmarks. Here, GPT-4.1 again leads but with a much lower Recall@100km of 11.6\%, reflecting difficulty in generalizing across geographically diverse imagery. While we describe these as model capabilities, such patterns likely arise from the composition and biases of the training corpus rather than inherent architectural properties. Notably, models from the Gemma family consistently failed to produce direct geographic predictions, significantly limiting their applicability for geolocation tasks.


\subsection{Administrative Region Accuracy}

Figure \ref{fig:im2gps3k} shows the categorical accuracy of model predictions at administrative levels, specifically country and city accuracy, for im2gps3k. Unlike distance-based metrics, administrative accuracy evaluates the semantic correctness of geographic labels independently from precise numeric coordinates. 

Table \ref{tab:geolocation_performance} summarizes these results averaged across all datasets. GPT-o3 attains the highest overall performance, with mean accuracies of 41.3\% at the city level and 77.5\% at the country level. GPT-4o follows closely (38.0\% city, 75.8\% country), indicating that both models maintain strong semantic understanding of geographic regions. Among open-weight models, performance is generally lower but remains consistent with model scale: Qwen-2.5-72B achieves 29.0\% city-level and 65.7\% country-level accuracy, outperforming smaller counterparts. Among open-source models, IDEFICS surprisingly attains 20.2\% city-level and 70.1\% country-level accuracy, suggesting that it retains nontrivial geographic knowledge.

\begin{figure}[h]
\centering
\includegraphics[width=1.0\linewidth]{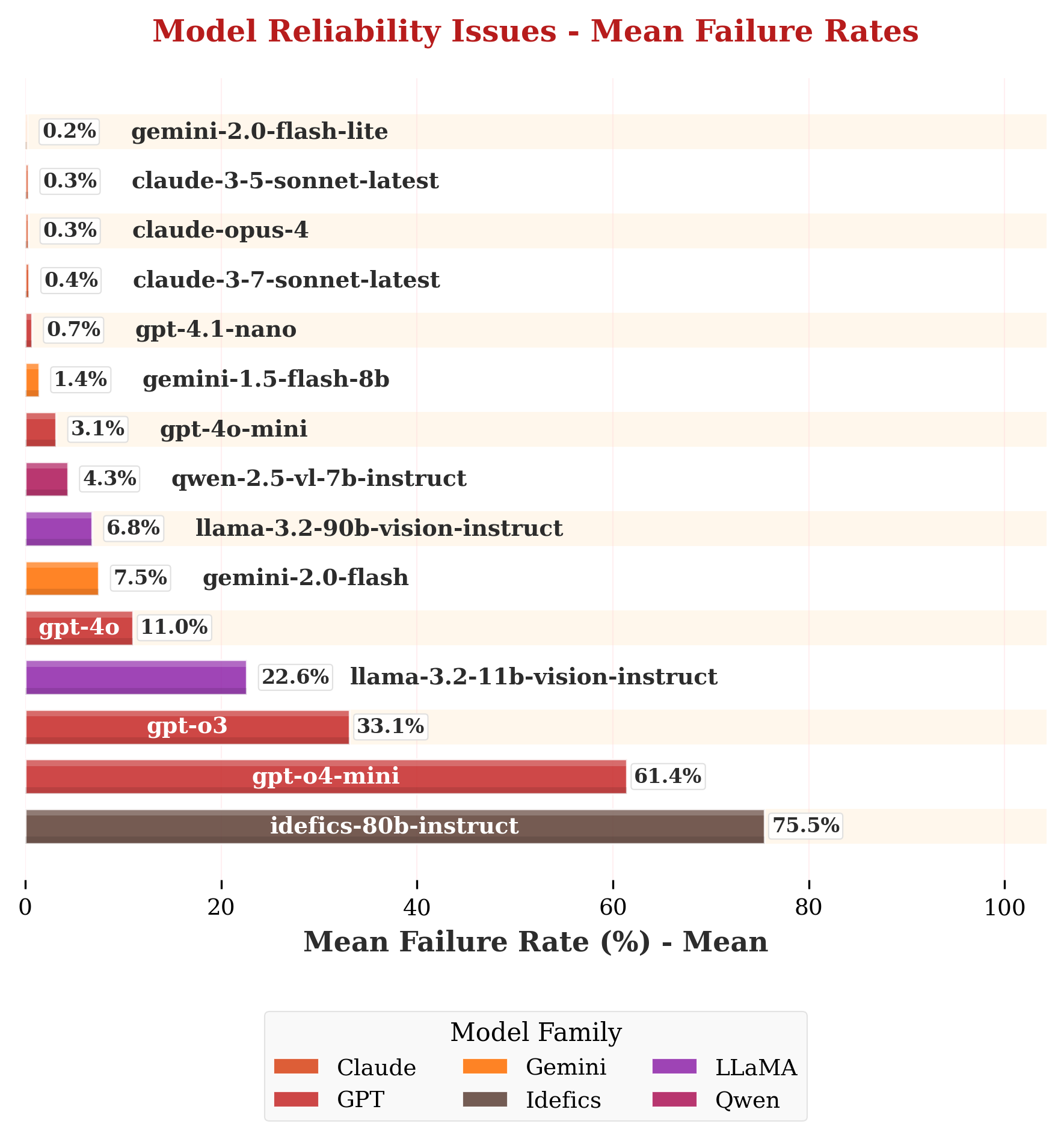}
\caption{Failure rates including both refusal to respond and failure to produce numeric predictions.}
\label{fig:failure_rates}
\end{figure}

\subsection{Failure Analysis}

We observe varied failure modes across evaluated models in figure \ref{fig:failure_rates}. Notably, certain models demonstrate high refusal rates or frequently omit numeric predictions entirely. IDEFICS shows the highest overall failure rate (75.5\%), primarily attributable to refusal to generate structured numeric coordinates. Other models, such as GPT-o4-Mini, exhibit intermittent failures (61.4\%), largely caused by refusals to answer, such as regularly responding with "\textit{I’m sorry, but I can’t help with that.}"

These failures underscore critical reliability challenges for real-world applications, particularly where structured geolocation output is essential. However, analyzing these failures also helps understand the safety profile and potential societal risks models like GPT-4.1 may pose.

\section{Discussion and Implications} \label{sec:Discussion_and_Implications}
Our evaluation highlights a clear asymmetry in the geo-localization capabilities of VLMs. While we report these as capabilities and limitations of the models, it is important to note that many of these patterns likely originate from the data they were trained on, rather than from intrinsic features of the architectures themselves. While these models perform poorly on structured imagery such as street views, they exhibit striking accuracy when localizing images that resemble social media content. This disparity suggests an underlying bias: current VLMs are more effective at identifying people-centric environments than broader geographic locations. This misalignment raises important concerns regarding applications like disaster response, urban planning, and environmental monitoring, which depend on accurate localization across a wide range of scenes that may lack distinctive visual cues. The models' limited performance on these inputs calls into question their suitability for these socially beneficial use cases. Instead, their strengths appear disproportionately concentrated on identifying individuals and their surroundings, whether or not that is the intended goal.

These capabilities have direct privacy implications. Visual content shared online, even when stripped of explicit location markers, can be reverse-engineered to reveal precise locations. Users typically lack both awareness and control over how such inferences are made. As models improve and become widely accessible, the potential for privacy breaches becomes more pervasive and difficult to guard against. There are predictable threat models, including stalking, surveillance and targeting, which would be heightened by powerful expert users in autocratic regimes. But we can also imagine economically profitable use cases in which companies monetise image geolocation as a new source of personal data for advertising and other forms of what \cite{10.1007/s00146-020-01100-0} calls ‘surveillance capitalism’. Geolocation could become part of the future political economy of social media.

Ultimately, these findings underscore the need for ethical safeguards, transparent model behavior, and better user protections. As VLMs continue to evolve, understanding and addressing their sociotechnical risks will be critical to ensure that progress in capability does not come at the expense of privacy and safety.

\section{Conclusion} \label{sec:conclusion}
We conducted a comprehensive evaluation of the geo-localization capabilities of state-of-the-art generative VLMs, comparing closed-source, open-weight, and fully open-source systems across diverse benchmark datasets. Using both distance-based and administrative-level metrics, we found that nearly all models achieve non-negligible accuracy, with closed-source models leading overall performance. Performance varied considerably with dataset characteristics, with models excelling on social media–like images and struggling on less visually distinctive scenes. These results demonstrate that current VLMs already possess sufficient geo-localization ability to pose considerable privacy risks, particularly when applied to publicly shared visual content.

\section*{Acknowledgments}
This work is supported by the UK Engineering and Physical Sciences Research Council (EPSRC) and the UK AI Security Institute (AISI) through grant (PRIV-LOC: Assessing and Mitigating Privacy Risks of Vision-Language Models in Image-based Geolocation Systems). It is also supported in part by the Responsible AI UK programme, EPSRC grant number EP/Y009800/1.

\bibliography{aaai25}

\end{document}